\documentclass[letterpaper, 10 pt, conference]{ieeeconf}  % Comment this line out if you need a4paper

\IEEEoverridecommandlockouts                              % This command is only needed if 
\usepackage{graphicx} % for pdf, bitmapped graphics files
\graphicspath{{images/}}
\usepackage{amsmath} % assumes amsmath package installed
\usepackage{algorithm}
\usepackage{algorithmic}
\usepackage{xcolor}
\usepackage{amsfonts,amssymb}
\usepackage{threeparttable}

\title{\LARGE \bf
Enhancing Grasping Performance of Novel Objects through an Improved Fine-Tuning Process
}

\author{Xiao Hu$^{1}$,Xiangsheng Chen$^{1,2}$
\thanks{$^{1}$Xiao Hu is with the School of Engineering, Shenzhen University,Shenzhen,China {\tt\small huxiao2021@email.szu.edu.cn}
The work is done when he is interning at Xpeng Robotics }.
\thanks{$^{2}$XiangSheng Chen ,Academician of the Chinese Academy of Engineering,
        Shenzhen,China
        {\tt\small xschen@szu.edu.cn}}%      
}

\begin{document}
\maketitle
\thispagestyle{empty}
\pagestyle{empty}

%%%%%%%%%%%%%%%%%%%%%%%%%%%%%%%%%%%%%%%%%%%%%%%%%%%%%%%%%%%%%%%%%%%%%%%%%%%%%%%%
\begin{abstract}

Grasping algorithms have evolved from planar depth grasping to utilizing point cloud information, allowing for application in a wider range of scenarios. However, data-driven grasps based on models trained on basic open-source datasets may not perform well on novel objects, which are often required in different scenarios, necessitating fine-tuning using new objects. The data driving these algorithms essentially corresponds to the closing region of the hand in 6D pose, and due to the uniqueness of 6D pose, synthetic annotation or real-machine annotation methods are typically employed. Acquiring large amounts of data with real-machine annotation is challenging, making synthetic annotation a common practice. However, obtaining annotated 6D pose data using conventional methods is extremely time-consuming. Therefore, we propose a method to quickly acquire data for novel objects, enabling more efficient fine-tuning. Our method primarily samples grasp orientations to generate and annotate grasps. Experimental results demonstrate that our fine-tuning process for a new object is 400 \% faster than other methods. Furthermore, we propose an optimized grasp annotation framework that accounts for the effects of the gripper closing, making the annotations more reasonable. Upon acceptance of this paper, we will release our algorithm as open-source.

\end{abstract}

%%%%%%%%%%%%%%%%%%%%%%%%%%%%%%%%%%%%%%%%%%%%%%%%%%%%%%%%%%%%%%%%%%%%%%%%%%%%%%%%
\section{INTRODUCTION}
\begin{figure}[ht]
        \centering
        \includegraphics[width=0.48\textwidth]{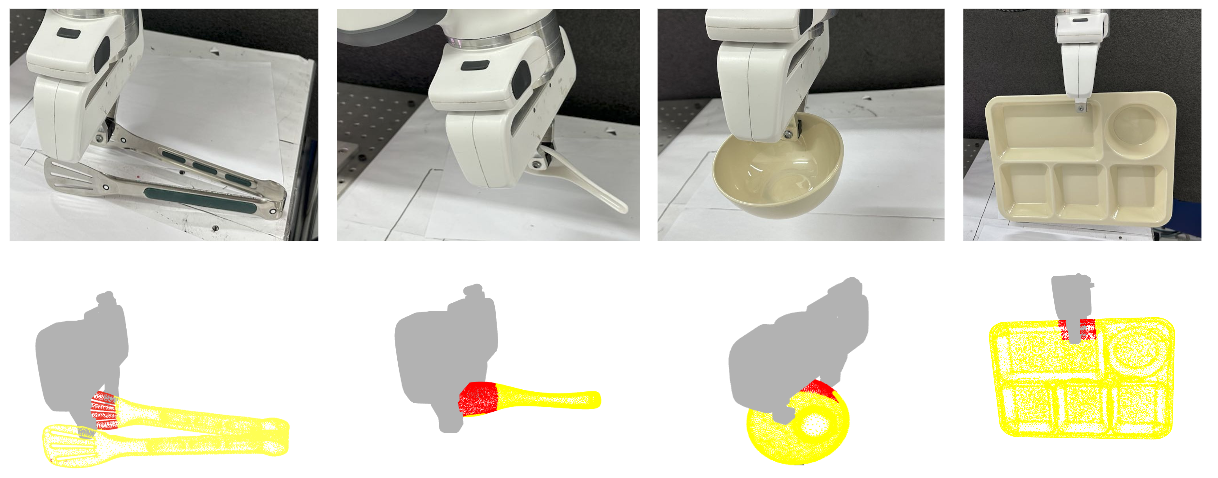} % 插入图片
        \caption{\textbf{The closing region of the hand} Generated by different objects varies, necessary for point cloud-based grasping algorithms.} % 添加图片标题
        \label{fig:intro} % 添加图片标签，方便引用
\end{figure}

%当前的抓取算法效果比较好的是基于点云的抓取算法，这类算法的核心是与抓取配置对应的夹爪闭合区域
Grasping has always been a crucial task in the field of robotic manipulation.
Due to the diversity of object shapes, robotic grasping faces a severe problem.
In recent years, data-driven methods have made impressive progress against the problem\cite{c6}.
%Current grasping algorithms\cite{c4}\cite{c5}\cite{c10} primarily focus on planar depth-based approaches, such as GGCNN\cite{c1}.
%However, due to their reliance on planar execution, these algorithms exhibit lower grasping success rates in dense clutter scenarios compared to point cloud-based grasping algorithms, such as PointnetGPD\cite{c2} and GraspNet\cite{c3}.
Representative point cloud-based algorithms, such as PointnetGPD\cite{c2} and GraspNet\cite{c3}, show distinct performance in dense clutter scenes.
The core aspect of those algorithms lies in the input data\cite{c7}, (see \textbf{Fig}. \ref{fig:intro}) called the closing region of the hand, which contains a portion of the object point cloud within the gripper.

% 这些基于点云的抓取算法流程是，off line阶段：对物体采样抓取并标注。on line阶段输入场景点云，采样抓取，输入每个抓取的夹爪闭合区域点云，输出这些抓取配置的分值，然后选择分值最高的执行
% Among the currently popular point cloud-based grasping algorithms\cite{c9}\cite{c12}, these methods are data-driven and primarily rely on generating data for the closing region of the hand using open-source datasets.
% Subsequently, force closure\cite{c13} (a method based on the force friction cone) is utilized for scoring and annotation.
Corresponding to the input data, labels used for training are the scores of force closure\cite{c13} (a method based on the force friction cone).
% The data is then trained to obtain a model.
During the grasping execution phase, heuristic grasping sampling is performed on the scene point cloud (input from the depth camera).
The resulting closing region of the hand for all samples is input into the network for inference, and the grasp with the highest score is selected for execution.
%对比其他抓取算法基于点云的6D pose抓取在不同的场景中具有更高的抓取成功率
Compared with other grasping algorithms\cite{c16}, point cloud-based grasping methods can perform grasping tasks more effectively.
This is mainly attributed to their output of 6D pose grasping information\cite{c14} (including grasping points and grasping orientation) and the adoption of basic network models for point cloud processing.
This endows point cloud-based grasping algorithms with a certain degree of generalization when executing grasping tasks.
Moreover, due to the richer 6D pose information output, the grasping success rate of these methods is higher than that of other techniques.

%基于点云的抓取算法的局限性在于，当抓取novel object的时候通常会抓取失败，所以我们需要对novel object生成抓取并fine tune模型
Although these grasping algorithms perform well, they may fail to grasp some novel objects successfully.
But in the numerous application scenarios of grasping tasks, there is often a demand for handling the challenge of grasping novel objects.
Therefore, we need to fine-tune the grasping dataset to achieve successful grasping of these novel objects.

% Due to the diversity of object shapes, robotic grasping poses challenges. In recent years, data-driven methods have made impressive progress in grasping. Current mainstream grasping algorithms, such as PointNetGPD and AnyGrasp, are data-driven. Hence, we need to create grasping datasets for some objects in real-world scenarios and perform fine-tuning to adapt these algorithms for grasping new objects.

% For these grasping algorithms, the core input during training is the closing region of the hand contains and its corresponding score, which are generated by heuristic grasping algorithms. These methods represent grasps using the center of two parallel jaws and the gripper's 6D Pose, with each grasp corresponding to the closing region of the hand contains on the object. To generate viable grasps, they need to sample and evaluate grasp candidates and score them using force closure.

%基于点云的抓取算法的抓取数据集生成都是通过基于Antipod的方式生成的
These methods generate grasps based on the contact points of parallel jaws on the object and produce grasps by performing collision detection and random sampling of gripper poses within a certain angular range. 

% This generation method is time consuming, typically requiring tens of thousands of points to represent a common object. On a mainstream computer configuration, sampling thousands of grasps for a novel object takes six to seven hours, which is obviously very time consuming. We propose that using predefined grasping vectors can generate a sufficient number and efficient grasps for a new object within an hour, requiring only the object's point cloud as input for the algorithm. This can greatly improve the efficiency of fine-tuning and promote the widespread use of these grasping algorithms.

\begin{figure*}[htbp]
        \center{\includegraphics[width=18cm]  {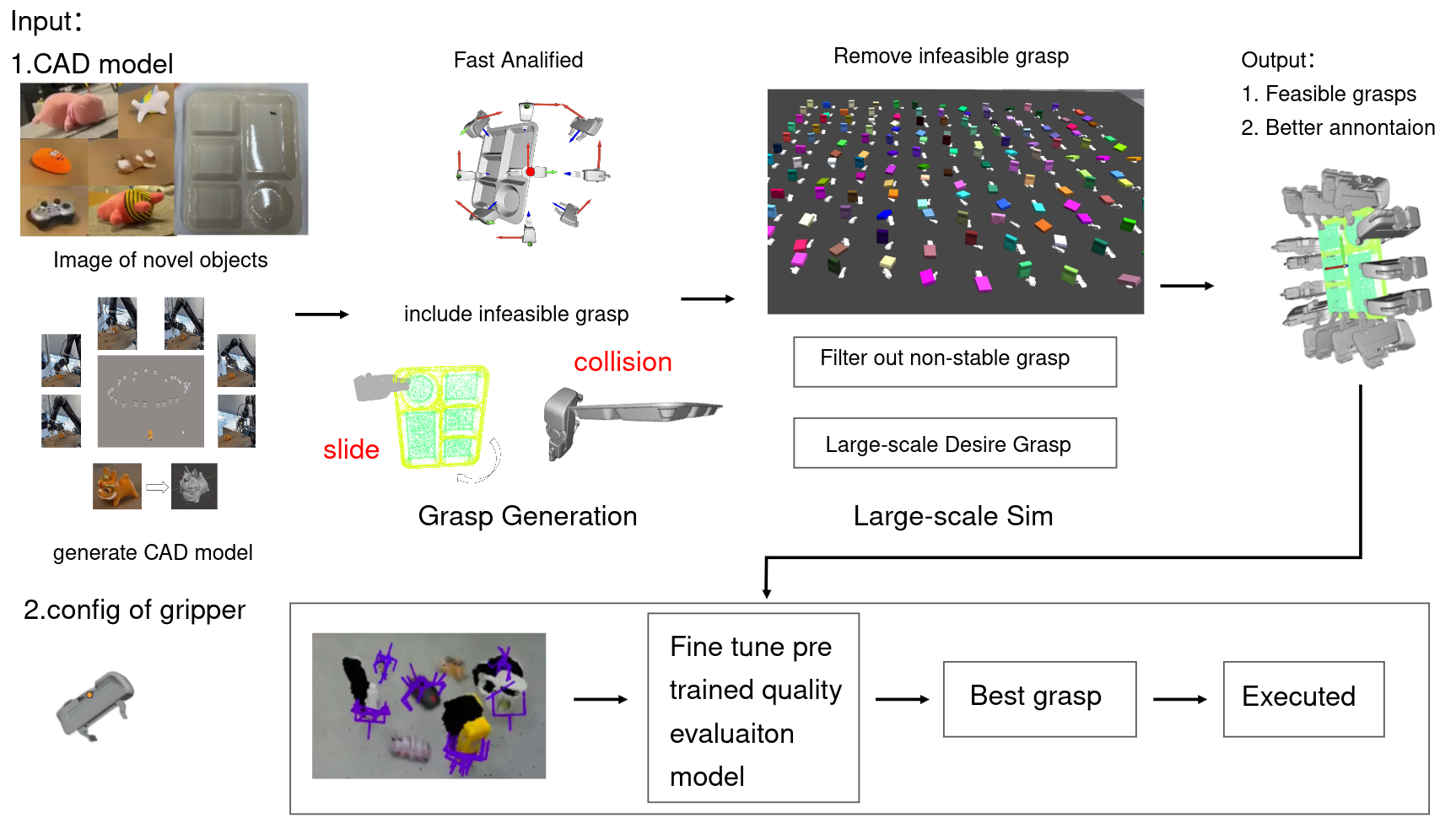}} 
        \caption{{\bf Framework}:Our workflow initially generates grasp configurations rapidly through geometric analysis, followed by the elimination of infeasible grasps via large-scale simulation, and ultimately outputs a rescored grasp dataset for fine-tuning.}
        \label{fig:framework}
\end{figure*}  
%生成抓取数据集的时候非常的慢，我们提出一种更快的方法生成抓取，大大加速了抓取数据集生成的过程
The core of these methods lies in the closing region of the hand  (see \textbf{Fig}. \ref{fig:intro}) in the dataset and their corresponding score annotations.
Therefore, we need to generate the corresponding closing region of the hand contains and score annotations for novel objects.
However, using the Antipod method\cite{c15} to generate grasps for novel objects is time-consuming on mainstream home computer configurations (Intel Core i7 13th \& RTX 3060), requiring 10 hours to generate 6500 grasps with annotation for one object.
Hence, we propose a faster grasp generation method, which takes only 1.2 hours to generate 6500 grasps with annotation for an object, and the grasp quality is comparable to the previous method.

% In addition, the aforementioned heuristic generation algorithms do not consider that when the gripper closes , the two sides of the gripper do not make contact with the object simultaneously. The side that makes contact first will cause the object to shift, as the closing region of the hand contains is not symmetrical relative to the center axis of the gripper. This situation leads to inaccurate scoring of grasping configurations generated by heuristic algorithms. To address this issue, we correct it through large-scale simulation, making grasp scoring more accurate and effective.

%而且Antipod的方式存在一些不合理的地方，比如存在一些抓取夹爪闭合的时候有一侧夹爪先碰到物体，导致物体发生移动从而抓取失败，除此之外，由于夹爪实际的提供的接触力有限，当抓取结束后抬起夹爪的时候由于力矩不平衡，物体发生转动导致静摩擦变为动摩擦，摩擦力减小最终物体掉落。
Moreover, the Antipod method relies on a pair of fixed points on the gripper as the contact points with the object to generate grasps.
However, during the execution of grasping, other points on one side of the gripper may touch the object first, causing displacement of the object.
This is due to the fact that the closing region of the hand contains is not symmetric with respect to the central axis of the gripper.
Consequently, this situation leads to inaccurate scores for the generated grasp configurations.
Additionally, some grasp configurations may cause the object to rotate and eventually fall due to an imbalance of torque when closing the gripper and lifting the object, as the contact force is limited by the gripper's power constraints.
To address these two issues, we make corrections in large-scale simulations, making the grasp scores more accurate and effective.

% \begin{figure*}[htbp]
%         \center{\includegraphics[width=18cm]  {images/overview.png}} 
%         \caption{{\bf Our Framework}:We extract features from multiple objects, filter out duplicate features using a feature selector to obtain a finite set of object point clouds in the gripper coordinate system, and then assemble them using a classifier.}
%         \label{fig:framework}
% \end{figure*}  

Our main contributions are:

\begin{itemize}
\item[$\bullet$]We propose a framework generation algorithm that is significantly faster than previous methods.
\item[$\bullet$]We introduce a filtering approach reduce duplicated and infeasible grasps.
\item[$\bullet$]Real-world experiments demonstrate the effectiveness of fine-tuning.

\end{itemize}

\section{Related Work}

% Over the years, grasp has been a popular research topic in robotics. Grasping annotations and datasets, as significant advancements in data-driven grasping, have introduced numerous annotation methods and datasets. Although some existing grasping datasets are abundant, they do not adequately meet the needs of current grasping algorithms.

Point cloud based grasping algorithms primarily rely on offline grasp generation methods to generate the closing region of the hand and their corresponding score annotations.
\subsection{Grasp Configuration Detection}

Given an object or a dense clutter scene, the goal of the grasping algorithm is to find and execute the optimal grasp at the moment. Current data-driven algorithms are mainly divided into two categories: planar depth grasping and point cloud-based 6D pose grasping. In planar depth grasping, leading algorithms in terms of grasping success rate include GGCNN\cite{c1}, GQCNN\cite{c27}, and others. The core idea of these algorithms is consistent. Taking GGCNN as an example, during the offline stage, a large number of RGB images of objects and their aligned planar depth maps are collected, and then these images are annotated with grasps (manually using rectangular bounding boxes). These data are then used to train the network. In the online stage, the input is the RGB image and its corresponding depth map, and the output is the grasp confidence, width, and angle for each pixel. The main drawback of these methods is the loss of the gripper's degrees of freedom, which can only perform grasps from above. Therefore, flat objects such as pizza boxes cannot be grasped from above. Moreover, these algorithms have a lower grasping success rate when dealing with dense clutter scenes, so point cloud-based grasping algorithms are needed.

\subsection{Point cloud based grasping algorithms}
Liang et al. proposed the PointNetGPD\cite{c2} algorithm to address the problem of selecting grasp configurations from point clouds. This method mainly consists of two stages: offline and online. In the offline stage, grasp datasets are generated from object point clouds in open-source datasets, specifically by sampling grasps for each object point cloud based on the Antipod method\cite{c15}, with each grasp corresponding to the closing region of the hand and its score annotation. Subsequently, a network model is designed based on PointNet\cite{c28}, which takes the closing region of the hand as input and outputs the corresponding score. In the online stage, the scene point cloud (provided by a depth camera) is input, and GPD\cite{c14} is used to sample grasps from the scene point cloud, obtaining candidate closing regions of the hand contains. Then the grasp with the highest score is executed. Firstly, when grasping novel objects different from those in the open-source dataset, the algorithm's grasping success rate is much lower than that for grasping objects similar to those in the open-source object dataset. Secondly, the algorithm's offline stage of generating the closing region of the hand for an object requires ten hours on a mainstream home computer (Intel Core i7 13th), which is too time-consuming. Lastly, the score annotation is based on the force closure method, which does not take into account that the object may move and cause grasping failure when the gripper closes.

Fang et al. proposed GraspNet\cite{c3} as a benchmark for large-scale grasp generation. GraspNet takes the scene point cloud as input, specifically, generating grasps for objects in open-source datasets using the PointNetGPD\cite{c2} method during the offline stage. Then, GraspNet arranges and combines these objects in the scene, removing grasps that collide with other objects, thereby generating scene point cloud grasps different from previous single-object grasps. This method has a higher grasping success rate in dense clutter scenarios, as it fully considers the effective closing region of the hand contains for objects stacked in the scene. However, its shortcomings are the same as those of PointNetGPD, namely, the large time consumption for grasp generation and the lack of consideration for the scoring annotation method during the grasping process.

\subsection{Grasp Quality Metrics}

To evaluate the quality of grasps, many analytical methods have physically analyzed the geometric shape of the gripper configuration and the objects used for evaluating grasp quality. Force closure\cite{c13} and GWS\cite{c17} analysis are two mainstream grasp quality metrics. The force closure method considers the friction between the gripper and the object, while GWS can handle frictionless situations. These two metrics are very versatile and have driven the progress of numerous grasp research directions. However, these methods do not take into account that during the execution of grasping, other points on the gripper may contact the object first, causing the object to move and leading to the failure of some initially high-quality grasps. Therefore, we considered this situation and optimized the scoring.

\subsection{Grasp Annotations}
Goldfeder et al. created the Columbia Grasp Database using GraspIt combined with form closure. It takes approximately 10 minutes to generate 15 form closure grasps, and the total computation time to build the grasp dataset on a server is about one month. This computational cost is evidently too high and not suitable for use during fine-tuning.

Lei et al. proposed a fast grasp generation method ``Fast grasping " This method requires downsampling of the object first, followed by PCA analysis to obtain the principal axes. Then, planar grasp annotations are performed on the object's cross-section along two orthogonal principal axes, which are then converted back to 3D grasps. Although this method is significantly faster than others, its drawback is the low diversity of grasps, as the grasp vectors are all along the PCA principal axes constructed by the object, making it unsuitable for fine-tuning as well.

% Pas et al. proposed GPD , which is widely used for 6D pose grasp annotation. This method involves traversing the object to obtain a pair of sampled contact points, sampling grasps based on these contact points within a set pose range, and calculating collision and force closure scores. This method can be used to generate a grasp dataset, but its significant drawback is that generating thousands of grasps for an object on a mainstream computer configuration takes nearly ten hours. Fine-tuning requires generating grasp datasets for more than one group of objects, and the computation time is still too long. Therefore, a better method is needed to generate grasps more quickly and effectively.

% Eppner et al. proposed "A Billion Ways to Grasp" which systematically outlines various grasping methods and introduces a simulation-based way of grasp generation. Firstly, compared to other methods in the literature, which mainly generate grasp vectors based on object surface normals, our approach is based on grasp orientations. This not only ensures faster generation speed than other methods but also results in higher grasp diversity compared to the surface normal-based methods. Additionally, the simulation-based grasp generation method in this work is time-consuming. Moreover, due to the simulation environment, some parts of the objects to be grasped are in contact with the ground, making it impossible to generate grasps and consequently reducing the richness of grasping.

\section{Methods}

Our proposed method (see \textbf{Fig}. \ref{fig:framework}) mainly aims to generate supplementary grasp datasets for novel objects more quickly and optimize the scoring annotations. Specifically, we first generate a CAD model for the novel object, and then based on spatial sampling grasp poses, we traverse the object, generating grasps and scoring annotations after filtering. Subsequently, we rescore the scoring annotations based on the object's movement during the grasping process using large-scale simulation. Furthermore, we set the maximum contact force for the gripper and filter out grasps with unbalanced torques.

\subsection{Generate CAD models of novel objects}

% \begin{figure}[ht]
%         \centering
%         \includegraphics[width=0.4\textwidth]{images/generate_obj.png} % 插入图片
%         \caption{{\bf Generate CAD model}:The robotic arm captures the RGB image of the object based on depth feedback to generate a CAD model.} % 添加图片标题
%         \label{fig:generate_boj} % 添加图片标签，方便引用
% \end{figure}

As data-driven grasping algorithms may fail to successfully grasp some novel objects when applied to different scenes, we need to fine-tune them to better grasp these novel objects.

The first requirement is the CAD model of novel objects. However, creating a detailed model of an object is very cumbersome. We have designed a framework that requires only a robotic arm and an RGB-D camera to output the CAD model of a new object for subsequent grasp generation.

% Due to the large size of CAD datasets and the relatively poor maintenance of open-source datasets, some everyday objects with significant grasping features are missing in the downloading process. Therefore, we use this method to supplement the missing everyday objects with significant grasping features to ensure the completeness and applicability of the CAD model dataset.

First, we use the robotic arm to capture images based on the depth returned by the camera(see \textbf{Fig}. \ref{fig:framework}). The robotic arm takes 32 effective photos of the object from different angles. These photos are then input into Nerf (Reconstructing 3D scenes from a set of sparse 2D images) to generate the object's CAD model. Finally, the generated object is upsampled to output a more detailed CAD model for subsequent grasp generation.

% We employ the proposed method to generate object-oriented CAD models for open-source datasets and compare their performance. The results indicate that the resolution and data size of the generated CAD models are very close to those of the CAD models in the open-source datasets.

In this way, we achieve the CAD model of our novel objects in the same quality of open-source datasets which set for grasp generation.

% \begin{table}
% \centering
% \begin{tabular}{|l|l|l|}
% \hline
% Objects           & Ycb resolution~ & Ours   \\
% \hline
% 003\_cracker\_box & 49905           & 38002  \\
% 024\_bowl         & 34977           & 42124  \\
% 056\_tennis\_ball & 4411            & 13872  \\
% 030\_fork         & 1205            & 1428   \\
% \hline
% \end{tabular}
% \end{table}

\subsection{Generate Grasp}

% Traditional methods for generating grasps are relatively time consuming, 

The process of generating grasp datasets using Antipod-based methods, such as Columbia Grasp\cite{c23} and PointnetGPD\cite{c2}, are time-consuming,with thousands of grasps taking up to ten hours(On 12th Gen core i7 12700H). Therefore, we propose a new method to generate grasps, which can fine-tune more quickly.

First, the grasp configuration is defined as , 
\begin{equation}
\begin{aligned}
& \text{G} = {[p,r]}\in \mathbb{R}^{6}\\
\end{aligned}
\end{equation}
$p=[x,y,z]\in \mathbb{R}^{3}$ represents grapspoint , $r=[r_{x},r_{y}r_{z}]\in \mathbb{R}^{3}$ represents the gripper's orientation. 

The Antipod-based methods are based on contact points to generate grasps. Specifically, it first finds a point on the object, samples grasps with the gripper width as the radius, and then samples grasp poses within a certain range based on the sampled contact points. The grasp poses are evaluated for collision and force closure scores. 
% The algorithm complexity of the sampling part in this method is , which obviously causes many repetitive grasp configurations for parallel jaws, such as , 
which obviously causes many repetitive grasp configurations for parallel jaws, such as 
\begin{equation}
\begin{aligned}
{G[\frac{n1+n2}{2},r]} = {G[\frac{n2+n1}{2},r]}\\
\end{aligned}
\end{equation}
n1 n2 are the contact points.

Since we aim to obtain diverse closing regions of the hand during the fine-tuning process and to accelerate the grasp generation, we propose a framework that differs from other similar approaches\cite{c8}. We sample a set of grasp orientations that are uniformly distributed in space. These grasp orientations are used to generate grasps by iterating through every point on the object.

In the space we set a grasp orientation and then we obtain the collection of grasp orientation r.

\begin{equation}
\begin{aligned}
{r_{i}} \in {[r_{1},r_{2},...,r_{n}]}\\
\end{aligned}
\end{equation}

% \begin{figure}[ht]
%         \centering
%         \includegraphics[width=0.35\textwidth]{images/grasp_matrix.png} % 插入图片
%         \caption{{\bf Grasp Orientation}:uniformly sample grasp orientation in the spherical space, and then traverse all points on the object after filtering to generate grasps.} % 添加图片标题
%         \label{fig: Grasp_Orientation} % 添加图片标签，方便引用
% \end{figure}

The algorithm for our entire grasp generation process is as follows:

\begin{algorithm}
    \caption{Generate Grasp}
    \begin{algorithmic}[1]
        \ENSURE{\parbox[t]{0.8\linewidth}{CAD model, Config of parallel jaws}}
        \REQUIRE{$G$[grasppoint , grapsconfig , grasp quality]}
        \STATE Define uniformly sampled $R_{i}$ grasp approach orientation
        \STATE $normal_{i}$Point cloud redirection
        \STATE Iterate over the $n$ points on the object
        \IF{ gripper no collision with objects}
        \STATE{$g_{i}\in G$}
        \ENDIF
        \STATE{Remove repeated closing region}
        \STATE{Computational force closure grasping}
    \end{algorithmic}
\end{algorithm}

In this process, we use collision detection and local reorientation of the object's normal vector to ensure that the generated grasps do not collide with the object. The algorithm complexity of our sampling process is $O(n)$(Compared to the Antipod-based algorithm, the algorithm complexity is $O(n^{2})$ ). And then we remove duplicate closing region of the hand contains.Finally, we employ the force closure scoring strategy same as PointnetGPD\cite{c2} and GraspNet\cite{c3}, and the results show that our method is 400\% faster. The score distribution and grasp success rate of our method are basically consistent with previous algorithms.

\begin{figure}[ht]
        \centering
        \includegraphics[width=0.2\textwidth]{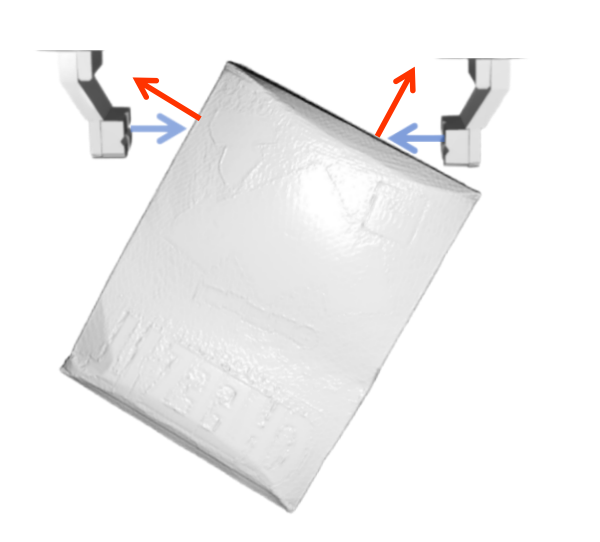} % 插入图片
        \caption{{\bf Remove bad grasp}:Grasps with contact force direction vectors exceeding the threshold are quickly filtered using surface vectors obtained from the differential point cloud of the object.} % 添加图片标题
        \label{fig: normal filter} % 添加图片标签，方便引用
\end{figure}

The reasons why our method is faster than the antipodal approach and achieves equivalent results are as follows. First, our method does not generate duplicate grasps, as previously mentioned, the contact point-based antipodal approach produces repetitive grasps. Second, we filter out unreasonable grasps based on the differential normal vectors of the object faces(see in \textbf{Fig}. \ref{fig: normal filter}), which is more time-efficient than using force closure to eliminate these grasps. Finally, we employ the same scoring and labeling method as the antipodal approach, ensuring consistency in the generated grasp data and scoring effectiveness.

\subsection{Stable grasp}

\begin{figure}[ht]
        \centering
        \includegraphics[width=0.35\textwidth]{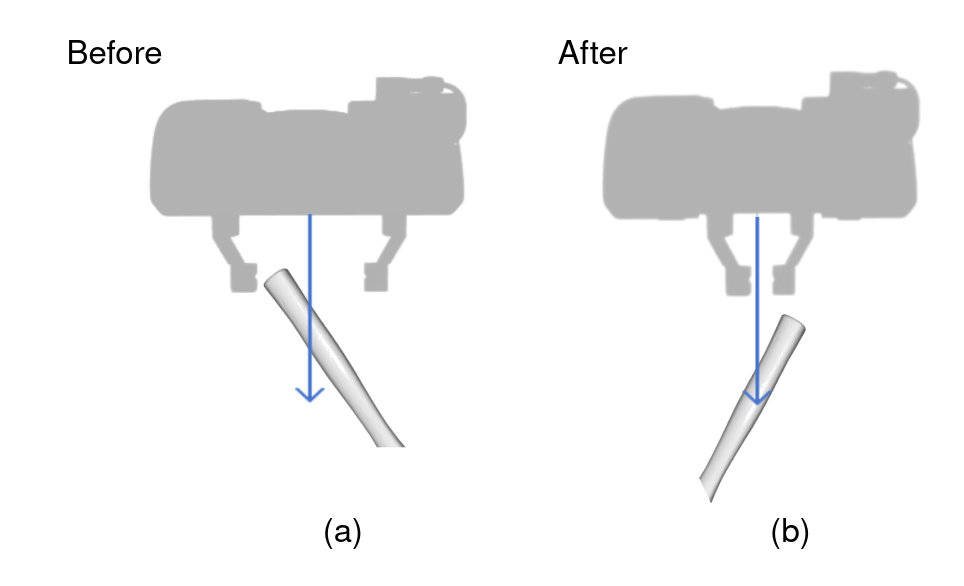} % 插入图片
        \caption{{\bf Statically generated grasps}:(a) shows the state before the gripper closes, in which the closing region of the hand is not symmetrical with respect to the grasp approach.(b) shows the state after the gripper closes, where the object fails to be grasped due to the displacement caused by the left gripper touching the object first.} % 添加图片标题
        \label{fig:stable closing region} % 添加图片标签，方便引用
\end{figure}

In the process of executing grasping, the gripper moves to the grasp point according to the grasp approach and then closes the gripper. However, during this process, an important step has been overlooked. As shown in \textbf{Fig}. \ref{fig:stable closing region}, when the gripper is closing, if one side of the gripper touches the object first, the force balance of the object will be disrupted, resulting in the object's movement.resulting in movement and an unstable grasp.

% \begin{figure}[ht]
%         \centering
%         \includegraphics[width=0.48\textwidth]{images/closingregion.png} % 插入图片
%         \caption{{\bf Statically generated grasps}:(a) In the figure, the grasp is stable, as the closing region of the hand contains exhibits a symmetrical configuration concerning the grasp approach, (b) Conversely, the grasp in the figure is unstable due to its asymmetrical configuration concerning the grasp approach.} % 添加图片标题
%         \label{fig:mold_penetration} % 添加图片标签，方便引用
% \end{figure}

% In the process of executing a grasp, the robotic arm moves to the grasp point according to the grasp approach and then closes the gripper. However, an important aspect has been overlooked in this process, which is that when closing the gripper, if one side of the gripper touches the object first, the object will break the force balance and cause movement. As shown in  , the left side of the gripper will touch the object first, 

We define a stable grasp as:

\begin{equation}
\begin{aligned}
& & PC_{1} \cdot r=PC_{2} \\
\end{aligned}
\end{equation}

$r=I-2uu^{T}\in \mathbb{R}^{3},u=[0,0,1]\in \mathbb{R}^{3}$ represents the mirror rotation matrix. 

$PC_{1} \& PC_{2} \in \mathbb{R}^{3}$ represent the point clouds on the left and right sides of the grasp approach within the closing region of the hand, respectively.

Slight differences may cause the gripper to touch the object on one side first during the grasping process, leading to object movement and grasping failure. Some grasps that are judged to be good under static force closure conditions may also fail during execution. Thus, we need to filter out these grasps. 

Our goal is to minimize the discrepancy between grasps generated under static conditions and the execution of these grasps.

\begin{equation}
\begin{aligned}
& Score(||S(p,r) - G(p,r)||) \\
\end{aligned}
\end{equation}

Optimize the grasps generated under static conditions $G(p,r)$ and the scoring annotations of grasps after the gripper closes $S(p,r)$ .

For the reasons mentioned above, we adopt large-scale simulation for grasp filtering. First, we perform the grasps generated under static conditions using large-scale simulation. After the gripper closes, if the object is outside the gripper space, the grasp is considered invalid and removed from the grasp configuration. In addition, after the grasp execution is completed, the score of the grasp is reassigned based on the ratio of the distance moved by the object's center of mass to the size of the gripper. This makes the annotation of the grasp dataset more effective.we have designed a simple weighting function.

\begin{equation}
\begin{aligned}
& & Weights=(1-\frac{d_{1}-d_{2}}{width})/2 \\
\end{aligned}
\end{equation}

d1 and d2 represent the Euclidean distances between the first points on both sides of the gripper to contact the object along the gripper closing direction and the object, respectively.
width reperesent the width of gripper.

\begin{figure}[ht]
        \centering
        \includegraphics[width=0.48\textwidth]{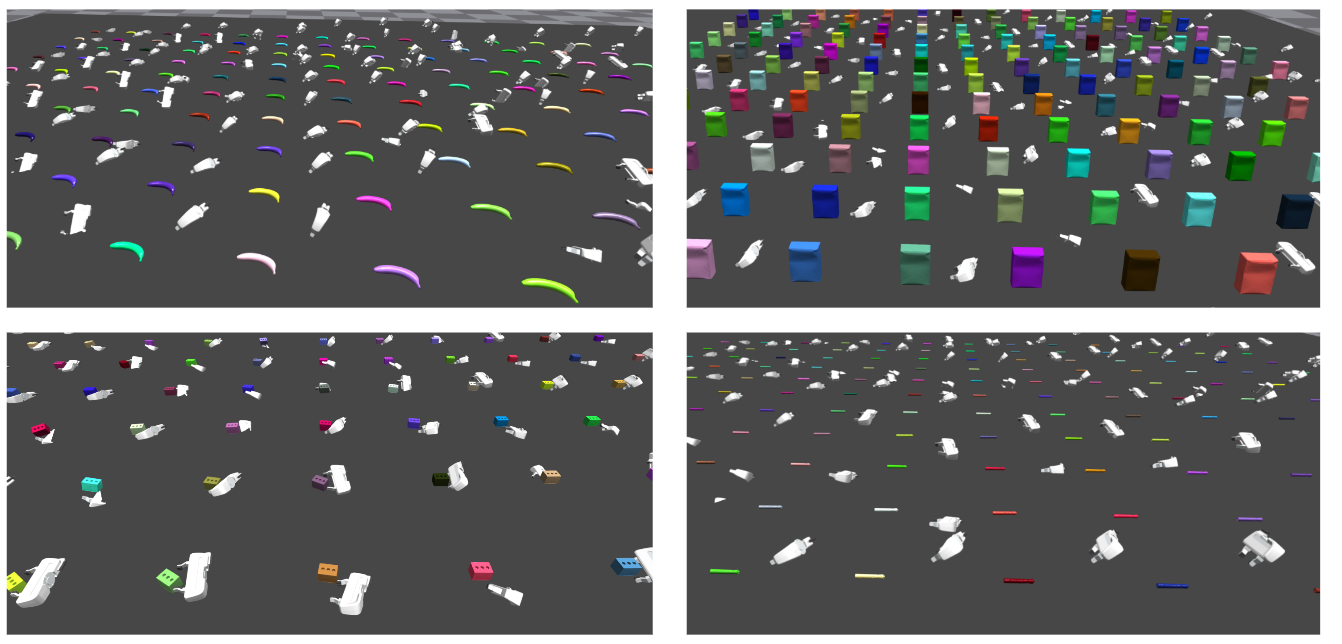} % 插入图片
        \caption{{\bf Screening}:Screening graspconfig Using Simulation\cite{c26}.} % 添加图片标题
        \label{fig:interaction} % 添加图片标签，方便引用
\end{figure}

 In some cases, grasp may fail, and objects may slip out of the gripper. This is mainly due to an upper limit on the contact force. When a larger torque balance is required for stable grasping, the grasp may fail.(bad grasp in \textbf{Fig}. ???\ref{fig:framework})

\begin{equation} \sum \vec{M} \ne 0 \end{equation}

% \begin{figure}[ht]
%         \centering        \includegraphics[width=0.35\textwidth]{images/torque_balance.png} % 插入图片
%         \caption{{\bf Torque Imbalance}:An imbalance of torque causes the object to rotate relative to the gripper, leading to the transition from static friction to dynamic friction, and ultimately resulting in the object slipping.} % 添加图片标题
%         \label{fig:Torque imbalance} % 添加图片标签，方便引用
% \end{figure}

To address the issue of power constraints and contact force limits during gripper closure. Firstly, we assign a certain mass to the objects to more accurately simulate the dynamic characteristics of objects during the actual grasping process. Next, we set the contact force limit during gripper closure to ensure that the gripper operation in the simulation environment conforms to the actual situation.

During the large-scale simulation, we test each grasp configuration and observe whether the object can be successfully moved to the target position after the gripper closes. We filter out grasp configurations that cannot successfully move the object within the contact force limit range. In this way, we can select grasp configurations that are more likely to succeed in actual operations, thereby generating better strategies.

\section{Experiment}
% \subsection{Necessity of Fine-tuning}
% Certain objects generate "closing region of the hand contains" that are not present in open-source grasping datasets. Therefore, fine-tuning is required for these objects in order to improve the generalization of the grasping algorithm. Since the actual input of the algorithm is the "closing region of the hand contains" from the dataset, we have compared the significant differences between the "closing region of the hand contains" of the new objects and those of other objects.

% \begin{figure}[ht]
%         \centering
%         \includegraphics[width=0.4\textwidth]{images/fpfh.png} % 插入图片
%         \caption{{\bf Difference}:31 objects selected from the YCB (Index 1-31)open-source dataset and  a novel object(Index 0) were used. The nearest neighbor distance was employed to compare the minimum distance between the closing region of the hand contains for each of the 31 objects and the novel object.} % 添加图片标题Index
%         \label{fig:differences} % 添加图片标签，方便引用
% \end{figure}

% \begin{equation}
% \begin{aligned}
% & & D=\sum||pc_{novel},pc_{base}||_{2} \\
% \end{aligned}
% \end{equation}

% By adopting the nearest neighbor method, the similarity is evaluated by comparing the difference D between the novel objects contained in the closing region of the hand and the base datasets.

\subsection{Generate Grasp}

% To compare the time consumption of our algorithm with that of other algorithms, we first employ a method to generate grasp points and grasp vectors, followed by annotating these grasps using force closure scoring. During the comparison, we ensure the consistency of experimental conditions, i.e., using the same computer hardware and output requirements. 

Our method significantly outperforms other approaches in terms of grasp generation speed. For the same set of novel objects, our method is capable of generating 6,500 optimal grasping configurations in just 1.2 hours, while other methods take an average of 10 hours to achieve the same results.(see in \textbf{Table} \ref{table:time cost})

% \begin{figure}[ht]
%         \centering
%         \includegraphics[width=0.48\textwidth]{images/time_cost.png} % 插入图片
%         \caption{{\bf Time Cost}:Our method significantly outperforms other approaches in terms of grasp generation speed. For the same set of novel objects, our method is capable of generating 6,500 optimal grasping configurations in just 1.2 hours, while other methods take an average of 10 hours to achieve the same results.} % 添加图片标题
%         \label{fig:interaction} % 添加图片标签，方便引用
% \end{figure}

% \begin{figure}[ht]
%         \centering
%         \includegraphics[width=0.48\textwidth]{images/scores.png} % 插入图片
%         \caption{{\bf Score Distribution}:The right-side image, which adopts force closure scoring, illustrates the distribution of grasp scores generated by our method, while the left-side image presents the distribution of traditional grasp scores generated based on contact points. Notably, both methods possess the same score distribution.} % 添加图片标题
%         \label{fig:interaction} % 添加图片标签，方便引用
% \end{figure}

We employ sampled grasp orientation to generate grasping actions. To evaluate and annotate these grasping actions, we adopt the force closure method, which is consistent with traditional methods. Throughout this process, we observe that the grasp score distributions generated by this approach remain similar for different novel objects, even though their shapes and characteristics vary. This indicates that our method is logically consistent and can be effectively applied to various types of objects. By combining the grasp orientation and force closure methods, we can ensure the logical clarity of the grasp generation process while maintaining its versatility for different objects.

\begin{table}
\caption{time cost}
\label{table:time cost}
\begin{threeparttable}
\centering
\begin{tabular}{l|l|l|l}
Methods               & Speed    & Quantity  & Quality    \\
\hline
PointnetGPD\cite{c2}           & 7.2s/per & 6500       & high       \\
Columbia grasp\cite{c23}        & 12s/per  & 5000       & high       \\
Fast Grasp\cite{c24}            & 0.5s/per & 2000       & low        \\
Billion ways to grasp\cite{c25}   & 240s/per & 15000      & very high  \\
Ours                  & 1.2s/per & 10000      & high      
\end{tabular}
\begin{tablenotes}
        \footnotesize
        \item[*] Speed represents the time taken to generate a grasp on the same computer configuration as previously described.%此处加入注释*信息
        \item[**] low indicates that the success rate of executing generated grasps is below 30\%, very high signifies a success rate of 95\% or higher, and high represents a success rate greater than 50\%." %此处加入注释**信息
      \end{tablenotes}
    \end{threeparttable}
\end{table}

\subsection{Large scale sim}

\subsubsection{Stable Grasp}

By utilizing large-scale simulation grasping, we are able to quickly filter out grasp configurations that may lead to failure when the gripper closes. Specifically, these unsuccessful grasp configurations are often due to the asymmetry of the closing region of the hand relative to the grasp approach, causing the gripper to contact the object first during the closing process and disrupt the object's equilibrium, resulting in object movement. This phenomenon is particularly evident in densely cluttered scenes, as unstable grasps can have a significant impact on the grasping performance in such scenarios.

% \begin{figure}[ht]
%         \centering
%         \includegraphics[width=0.48\textwidth]{images/empower.png} % 插入图片
%         \caption{{\bf Filter}:as shown in Fig From these figures, it can be observed that the original force-closure-based scoring method did not fully consider the potential relative movement of objects during grasping. To address this issue and make the scoring annotation more reasonable, we introduced a threshold to optimize the score distribution. Through this optimization approach, we can more effectively evaluate the feasibility of grasping and improve the success rate of grasping.} % 添加图片标题
%         \label{fig:interaction} % 添加图片标签，方便引用
% \end{figure}

In order to filter out effective grasp configurations, we employed a large-scale simulation approach. Specifically, we first created n simulation environments, in which the objects were placed in the same position and orientation. Subsequently, we performed grasping operations in each environment and observed the movement of the objects after the gripper closed. If the movement distance or orientation change of the objects exceeded the threshold, we considered the grasp configuration to be invalid and removed it from the candidate set. For those valid grasp configurations within the threshold range, we assigned weights to their annotated scores based on the object's movement distance and orientation change.

\begin{figure}[ht]
        \centering
        \includegraphics[width=0.48\textwidth]{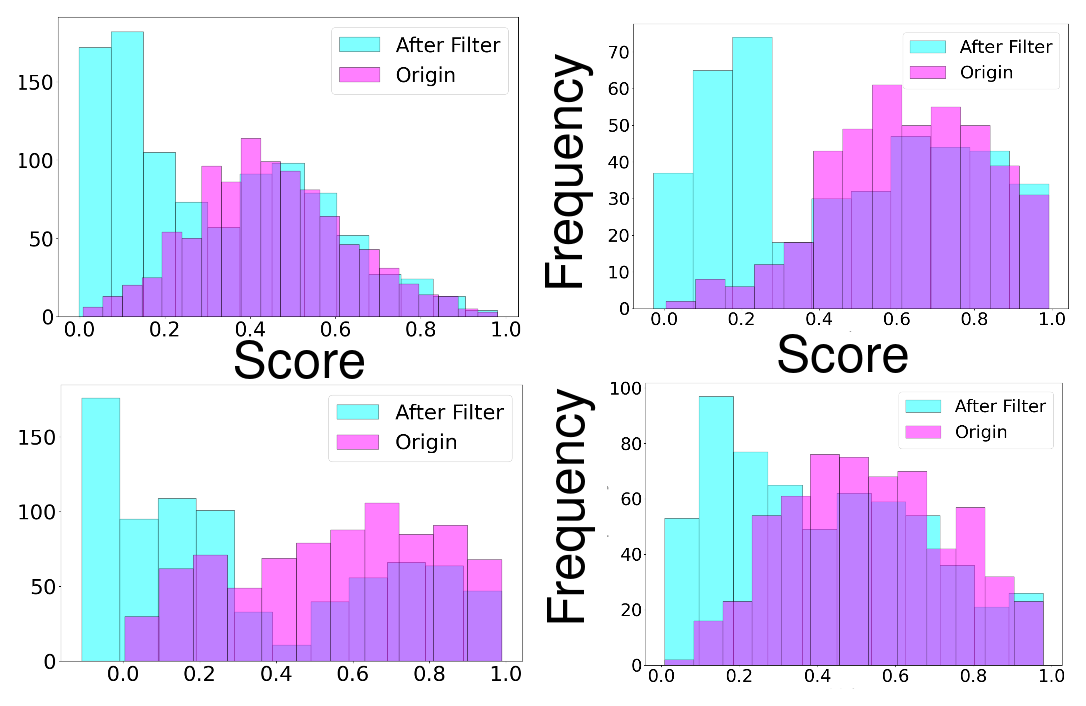} % 插入图片
        \caption{{\bf Rescore}:Four objects were selected, where the purple scores represent the original annotations based on force closure, and the cyan scores indicate the re-evaluated values. As shown in the figure, a large number of high-scored grasps in the original assessment would not perform well in practice due to collisions.} % 添加图片标题
        \label{fig:empower} % 添加图片标签，方便引用
\end{figure}

As shown in \textbf{Fig}. \ref{fig:empower}, it can be observed that the original force-closure-based scoring method did not fully consider the potential relative movement of objects during grasping. To address this issue and make the scoring annotation more reasonable, we introduced a threshold to optimize the score distribution. Through this optimization approach, we can more effectively evaluate the feasibility of grasping and improve the success rate of grasping(see in In the \textbf{table}. \ref{table:finetune1}).

\subsubsection{Desire Grasp}

In this filtering step, we primarily eliminate grasp configurations that, after the gripper closes, may result in torque imbalance due to limited gripper contact force. Specifically, when the gripper closes, if the contact force cannot generate sufficient torque to maintain the object's stability, the object may slide. At this point, static friction transitions to dynamic friction, reducing the frictional force and further exacerbating the object's sliding. Ultimately, this unstable grasping may cause the object to fall from the gripper. By filtering out these potentially failing configurations, we can optimize the grasping strategy.

% \begin{figure}[ht]
%         \centering
%         \includegraphics[width=0.48\textwidth]{images/piechart.png} % 插入图片
%         \caption{{\bf Gasping screening}:(a) illustrates the grasping screening process where the object's longest bounding line is greater than the gripper width, while (b) demonstrates the grasping screening process in which the longest bounding line is less than the maximum gripper width.} % 添加图片标题
%         \label{fig:Gasping screening} % 添加图片标签，方便引用
% \end{figure}

In our large-scale simulation setting with a maximum contact force of 25 N, we tested the antipodal method on a sample of 13 objects from the YCB dataset, each with 6,500 grasp configurations. We found that 42.2\% (35,659/84,500) of the grasps were ultimately unsuccessful after the gripper closure. By removing these failed grasps and fine-tuning the model training, we were able to improve the grasp success rate.

\subsubsection{Real robot experiments}

In our real-world experiments, the gripper employed is not a Franka model; however, it is actuated following the Franka's approach and also adopts a two-finger configuration."

\begin{table}
\centering
\caption{Real-world experiments}
\label{table:finetune1}
\begin{threeparttable}
\begin{tabular}{|l|l|l|}
\hline
Objects          & before finetune & after finetune  \\
            \hline
gamepad     & 0/3             & 3/3             \\
\hline
glue bottle & 3/3             & 3/3             \\
\hline
Tape        & 1/3             & 1/3             \\
\hline
Box         & 2/3             & 3/3             \\
\hline
doll\_01    & 1/3             & 1/3             \\
\hline
doll\_02    & 0/3             & 1/3             \\
\hline
doll\_03    & 2/3             & 2/3             \\
\hline
milk box    & 3/3             & 3/3             \\
\hline
Coke can    & 3/3             & 3/3         \\
\hline
\end{tabular}
\begin{tablenotes}
        \footnotesize
        \item[*] 1/3 indicates one successful grasp out of three trials. The above figure illustrates the results of the first fine-tuning process for the gamepad controller. doll see in \textbf{Fig}. \ref{fig:real_test} %此处加入注释*信息
      \end{tablenotes}
    \end{threeparttable}

\end{table}

In the \textbf{table}. \ref{table:finetune1} Prior to the fine-tuning of the game controller, all attempts to grasp it were unsuccessful. However, after fine-tuning using our proposed framework, the controller could be grasped effectively. Moreover, we conducted grasping tests on other objects, and the fine-tuning did not result in a decreased success rate for grasping these objects.

In addition, we conducted another set of real-world experiments, where three dolls, a game controller, and a milk carton were fine-tuned using our proposed workflow. The grasp success rate for each of the three dolls increased to 3/3, while the success rate for the other objects remained unchanged(see in \textbf{table}. \ref{table:finetune1}). These two sets of experiments demonstrate that our proposed framework can successfully grasp novel objects.

\begin{figure}[ht]
        \centering
        \includegraphics[width=0.48\textwidth]{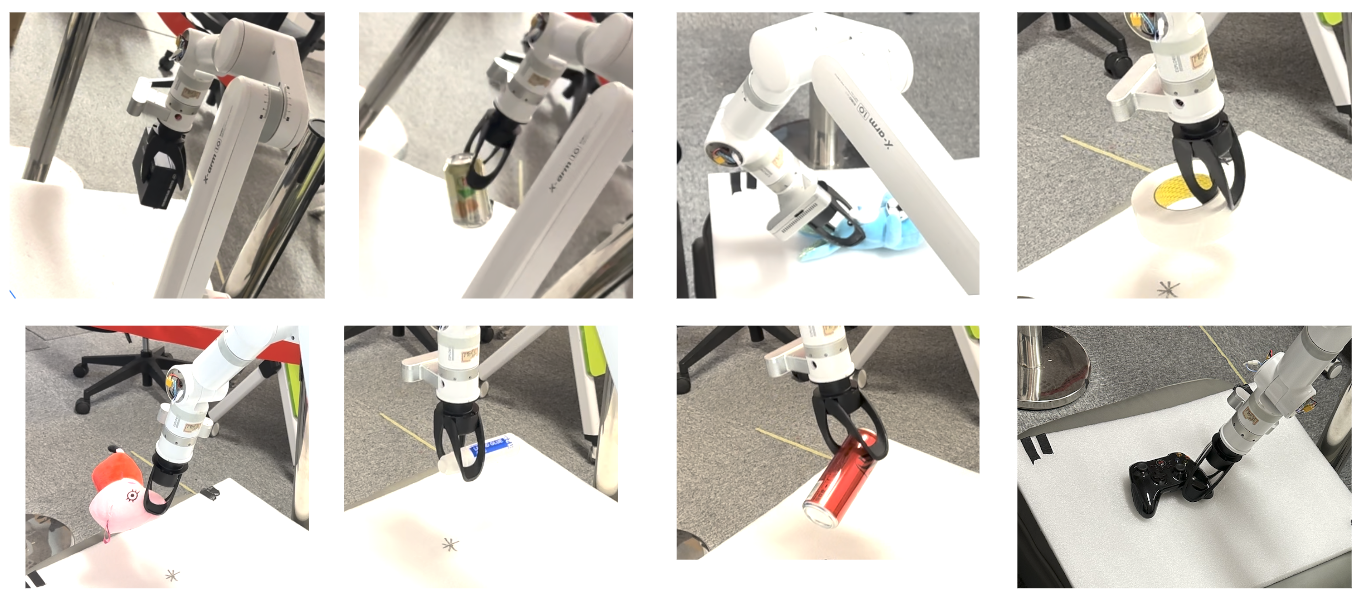} % 插入图片
        \caption{{\bf Real Test}:After fine-tuning, the model was tested on real robot grasping experiments using the objects employed during the fine-tuning process.} % 添加图片标题
        \label{fig:real_test} % 添加图片标签，方便引用
\end{figure}

\section{CONCLUSIONS}

Our proposed grasp generation method significantly reduces the time required for fine-tuning on novel objects. Furthermore, we introduce an optimized framework for annotating grasp scores, which improves the grasp success rate and provides more reasonable grasp score annotations. Ultimately, through real-world experiments, we demonstrate the effectiveness of our framework for fine-tuning on novel objects.

\addtolength{\textheight}{-1cm}   % This command serves to balance the column lengths

%%%%%%%%%%%%%%%%%%%%%%%%%%%%%%%%%%%%%%%%%%%%%%%%%%%%%%%%%%%%%%%%%%%%%%%%%%%%%%%%

%%%%%%%%%%%%%%%%%%%%%%%%%%%%%%%%%%%%%%%%%%%%%%%%%%%%%%%%%%%%%%%%%%%%%%%%%%%%%%%%

%%%%%%%%%%%%%%%%%%%%%%%%%%%%%%%%%%%%%%%%%%%%%%%%%%%%%%%%%%%%%%%%%%%%%%%%%%%%%%%%

\section*{ACKNOWLEDGMENT}

We would like to express our gratitude to BiDan Huang and JunKun Peng for their valuable suggestions on this article.

%%%%%%%%%%%%%%%%%%%%%%%%%%%%%%%%%%%%%%%%%%%%%%%%%%%%%%%%%%%%%%%%%%%%%%%%%%%%%%%%

\end{document}